\documentclass[Afour,sageh,times]{sagej}

\usepackage{moreverb,url}

\usepackage{hyperref}
\usepackage{graphicx}
\usepackage{mathptmx}
\usepackage{xcolor}
\usepackage{amsmath}
\usepackage{soul}

\newcommand{\code}{\texttt}

\newcommand{\add}{\textcolor{black}}

\newcommand\BibTeX{{\rmfamily B\kern-.05em \textsc{i\kern-.025em b}\kern-.08em
		T\kern-.1667em\lower.7ex\hbox{E}\kern-.125emX}}

\newcommand{\sd}{\cite{pham2009}}
\newcommand{\mmac}{\cite{torre2009}}
\newcommand{\gaze}{\cite{fathi2012}} 
\newcommand{\mpiic}{\cite{rohrbach2012}}

\newcommand{\salad}{\cite{stein2013}}
\newcommand{\ace}{\cite{shimada2013}}
\newcommand{\yc}{\cite{das2013}}
\newcommand{\kthc}{\cite{pieropan2014}}
\newcommand{\bb}{\cite{kuehne2014}}
\newcommand{\tumk}{\cite{tenorth2009}}

\newcommand{\opp}{\cite{roggen2010}}

\newcommand{\wwadl}{\cite{bruno2014}}

\newcommand{\gtea}{\cite{fathi2011}}
\newcommand{\mk}{\cite{pham2016}}
\newcommand{\uci}{\cite{rogez2014}}

\begin{document}
	
	\runninghead{Huang and Sun}
	
	\title{A Dataset of Daily Interactive Manipulation}
	
	\author{Yongqiang Huang and Yu Sun}
	
	\affiliation{The authors are with the department of Computer Science and Engineering, University of South Florida, Tampa, USA.}
	
	\corrauth{Yu Sun, Computer Science and Engineering, University of South Florida, 4202 E Fowler Ave., ENB 331,
		Tampa, FL 33620}
	
	\email{yusun@cse.usf.edu}
	
	\begin{abstract}
		Robots that succeed in factories stumble to complete the simplest daily task humans take for granted, for the change of environment makes the task exceedingly difficult. Aiming to teach robot perform daily interactive manipulation in a changing environment using human demonstrations, we collected our own data of interactive manipulation. The dataset focuses on position, orientation, force, and torque of objects manipulated in daily tasks. The dataset includes  \add{1,593} trials of 32 types of daily motions and 1,596 trials of pouring alone, as well as helper code. We present our dataset to facilitate the research on task-oriented interactive manipulation.
	\end{abstract}
	
	
	\maketitle	
	
	\section{Introduction}

	Robots excel in manufacturing which requires repetitive motion with little fluctuation between trials. In contrast, humans rarely complete any daily task by repeating \emph{exactly} what was done last time, for the environment might have changed. We aim to teach robots daily manipulation tasks using human demonstrations so that they are able to fulfill them in a changing environment. To learn how human finish a task by manipulating an object and interact with the environment, we need 3-dimensional motion data of the objects involved in fine manipulation motion, and data that represent the interaction.  
	
	Most of the currently available motion data are in the form of vision, i.e., RGB videos and depth sequences (for example, \gaze{}, \mpiic{}, \ace, \yc, \bb, \gtea, \uci), which are of little or no direct use to our purpose. Nevertheless, certain datasets exist which do include motion data. Slice \& Dice dataset \sd{} includes 3-axis acceleration of cooking utensils which are used while salads and sandwiches are prepared. 50 Salad dataset \salad{} includes 3-axis acceleration of more cooking utensils than Slice \& Dice which are involved in salad preparation. CMU-MMAC \mmac{} dataset includes motion capture and 6-degree of freedom (DoF) inertia measurement unit (IMU) data of the human subjects while the subjects are making dishes. The IMUs record acceleration in ($x, y, z$, yaw, pitch, roll). The Actions-of-Making-Cereal \kthc{} dataset includes 6-DoF pose trajectories of the objects involved in cereal making that are estimated from RGB-D videos. The TUM Kitchen dataset \tumk{} includes motion capture data of the human subjects while the subjects are setting tables. The OPPORTUNITY dataset \opp{} includes 3-D acceleration and 2-D rotational velocity of objects. The Wrist-Worn-Accelerometer \wwadl{} dataset includes 3-axis acceleration of the wrist while the subject is doing daily activities. The Kinodynamic dataset \mk{} includes mass, inertia, linear and angular acceleration, angular velocity, and orientation of the objects, but the manipulation exists in its own rights and does not serve to finish a task. 
	
	The aforementioned datasets are less than ideal in that 1) calculating the position trajectory using the  acceleration may be inaccurate due to accumulated error, 2) the motions of objects are not always emphasized or even available, and 3) all the activities are not fine manipulations that serve to finish tasks. Having identified those deficiencies, we collected a dataset ourselves that includes 3-dimensional ``position and orientation, force and torque" data of tools/objects being manipulated to fulfill certain tasks.  \add{The dataset is potentially suitable for learning either motion \cite{huang2015generating} or force \cite{lin2012learning} from demonstration, motion recognition  \cite{Subramani2017, Aronson?2016?5619} and understanding \cite{doi:10.1177/0278364911410459, Paulius2018, FLANAGAN2006650, SOECHTING2008565}, and is potentially beneficial to grasp research \cite{lin2016task, lin2015grasp, lin2015task, sun2016robotic}}.

	\section{Overview}
	
	We present a dataset of daily interactive manipulation. Specifically, we record daily performed fine motion in which an object is manipulated to interact with another object. We refer to the person who executes the motion as \emph{subject}, the manipulated object as \emph{tool}, and the interactive object as \emph{object}. We focus on recording the motion of the tool. In some cases, we also record the motion of the object.     
	
	The dataset consists of two parts. The first part contains \add{1,593} trials that cover 32 types of motions. We choose fine motions that people commonly perform in daily life which involve interaction with a variety of objects.  \add{We reviewed existing motion-related datasets \cite{huangbigdata, huang2016datasets, bianchi2016latest} to help us decide which motions to collect}. 
	
	The second part contains the pouring motion alone. We collect it to help with motion generalization to different environments. We chose pouring because 1) pouring is found to be the second frequently executed motion in cooking, right after pick-and-place \cite{pauliusfunctional} and 2) we can vary the environment setup of the pouring motion easily by switching different materials, cups, and containers. The pouring data contain 1,596 trials of pouring 3 materials from 6 cups into 10 containers.  
	
	We collect the two parts of the data using the same system. We specifically describe the pouring data in Sec. \ref{sec-pouring}.
	
	The dataset aims to provide position and orientation (PO) and force and torque (FT), nevertheless, it also provides \add{RGB and} depth vision with a smaller coverage. Table \ref{fig-motion_modalities} shows the number of trials and the counts of each modality for each motion. The minimum number of trials for each motion is \add{25}. Table \ref{tab-coverage-whole} shows the coverage of each modality throughout the entire data, where the coverage has a range of (0, 1], and a coverage of 1 means the modality is available for \emph{every} trial. The lower coverage of the vision modality is due to filming permission restriction. 
	
	\begin{table}
		\caption{The count for each modality for each motion. Each motion is coded \code{\textcolor{red}{m}x}, where \code{x} is an integer.}
		\label{fig-motion_modalities}
		\includegraphics[width=\linewidth]{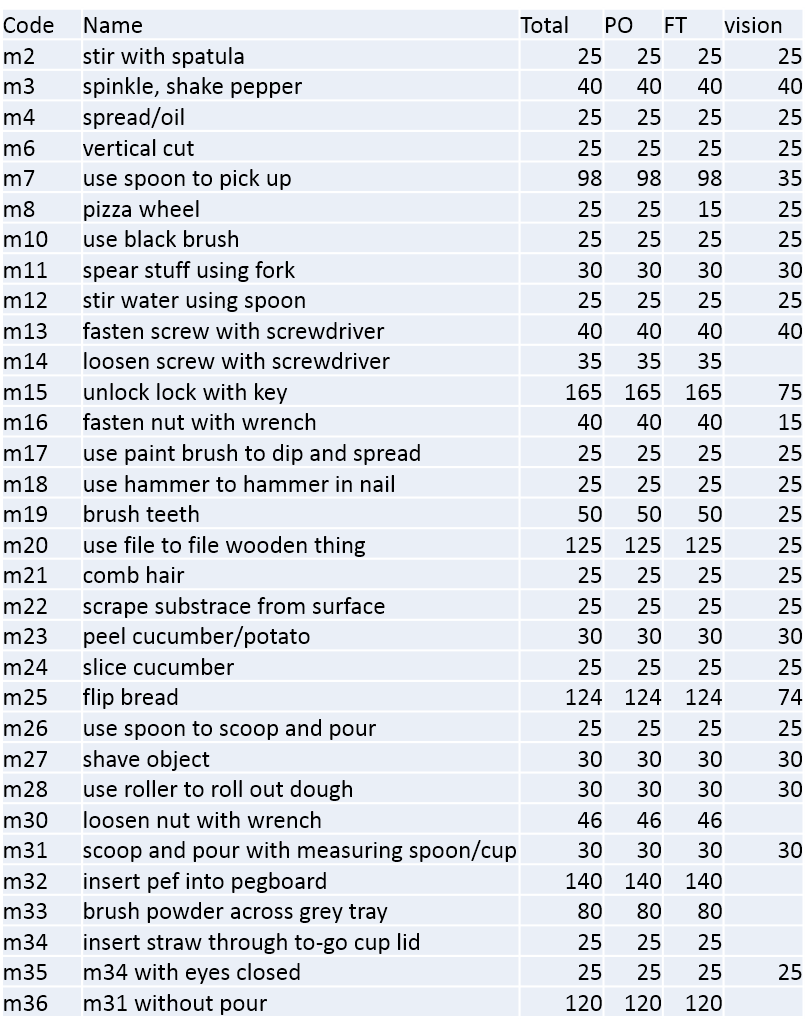}
	\end{table}
	
	\begin{table}[h!]
		\centering
		\caption{Modality coverage throughout the entire data.}
		\label{tab-coverage-whole}
		\begin{tabular}{ l c c c c c }
			\hline
			\textbf{Modality} & PO & FT & \add{vision} \\
			\hline
			\textbf{Coverage} & 1.0  & \add{1.0} & \add{0.50} \\
			\hline 
		\end{tabular}
		
	\end{table}
	
	\section{Hardware} \label{sec-hardware}

	\add{On a desk surface, we use blue masking take to enclose a rectangular area which we refer to as the working area, and within which we perform all the motions. We make a PrimeSense RGB+depth camera aim at the working area from above.}

\begin{figure}
		\includegraphics[width=\linewidth]{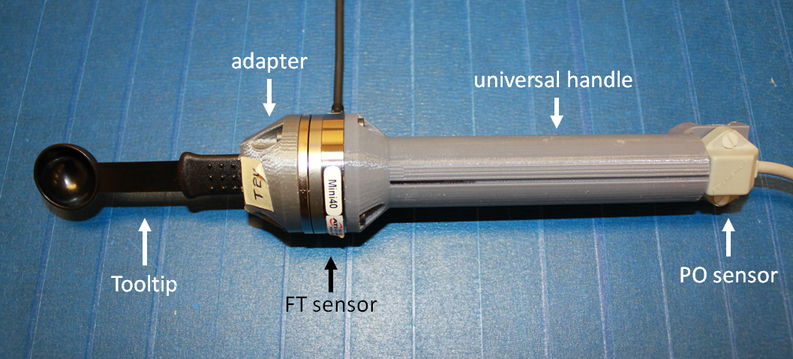}
		\caption{The structure that connects the tool, the FT sensor and the PO sensor}
		\label{fig-cascade}
	\end{figure}

\begin{figure}
		\centering
		\includegraphics[width=0.8\linewidth]{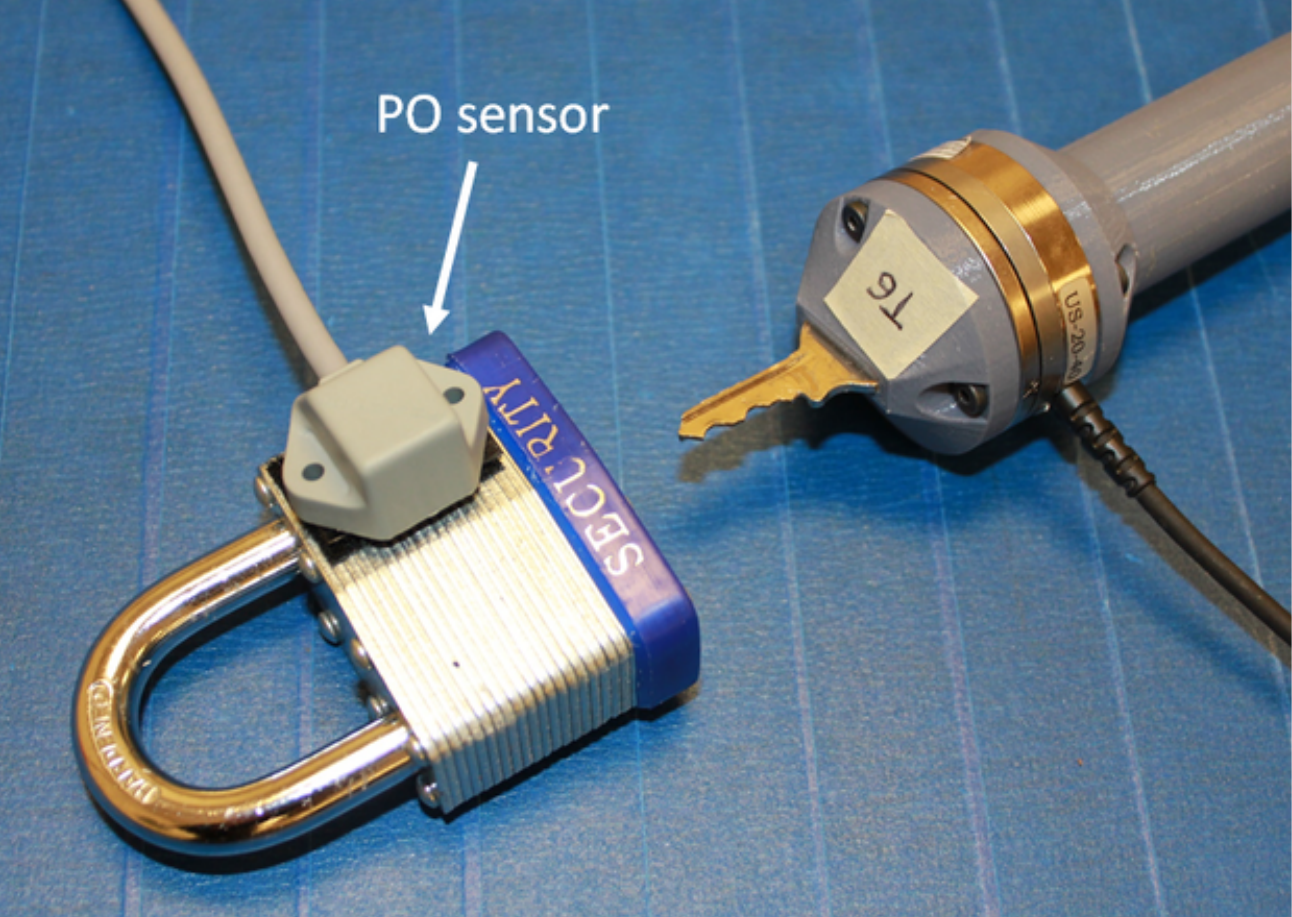}
		\caption{Tracking both the tool and the object with two PO sensors}
		\label{fig-tracking_object}
	\end{figure}

	\add{We started collecting PO data using the OptiTrack motion capture (mocap) system and soon afterwards replaced OptiTrack with the Patriot mocap system. Both systems provide 3-dimensional PO data regardless of their difference in technology.} Patriot includes a source and a sensor. The source provides the reference frame, with respect to which the PO of the sensor is calculated. 
	\add{We use an ATI Mini40 force and torque (FT) sensor together with the Patriot PO sensor.}
	 To attach both the FT sensor and the PO sensor to a  tool, we use a cascading structure that can be represented as: (tooltip + adapter + FT sensor + universal handle + PO sensor), where ``+" means ``connect". The end result is shown in Fig. \ref{fig-cascade}. A tool in general consists of a tooltip and a handle. We disconnect the tooltip from the stock handle, insert the tooltip into a 3D-printed adapter, and glue them together. Then we connect the adapter with the tooling side of the FT sensor using screws. We 3D-print a universal handle and connect it with the mounting side of the FT sensor using screws. At the end of the universal handle we mount the PO sensor using screws. In some cases, we track the object in addition to the tool, and to do that we put a second PO sensor on the object, as shown in Fig. \ref{fig-tracking_object}
	
\begin{figure}
		\includegraphics[width=\linewidth]{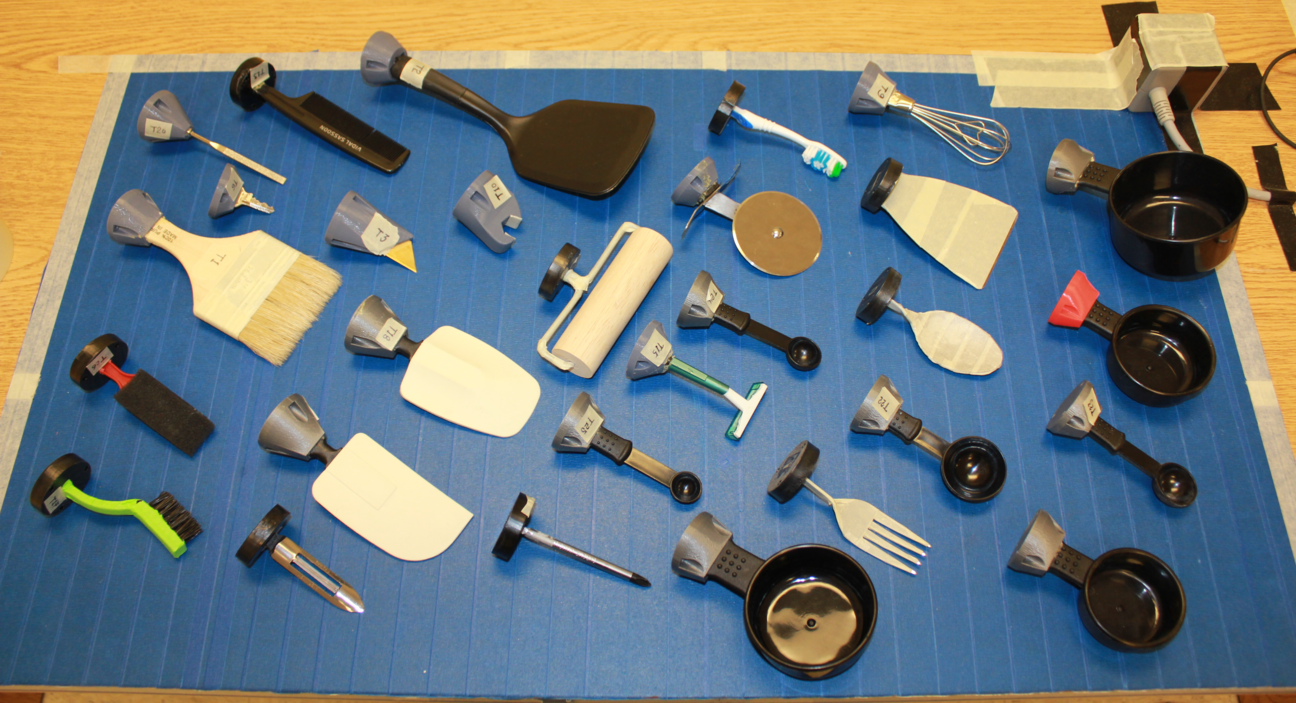}
		\caption{Examples of the tools that we have adapted}
		\label{fig-tools}
	\end{figure}

	Each tooltip is provided with a separate adapter. Since the tooltip and the adapter is glued together, a tool is equivalent to ``tooltip + adapter". Fig. \ref{fig-tools} shows the tools that we have adapted. 
	
	\section{Coordinate frames} 
	
	\newtheorem{definition}{Definition}
	
	\begin{figure}
		\centering
		\includegraphics[width=\linewidth, trim={6cm 2cm 5cm 2cm}, clip]{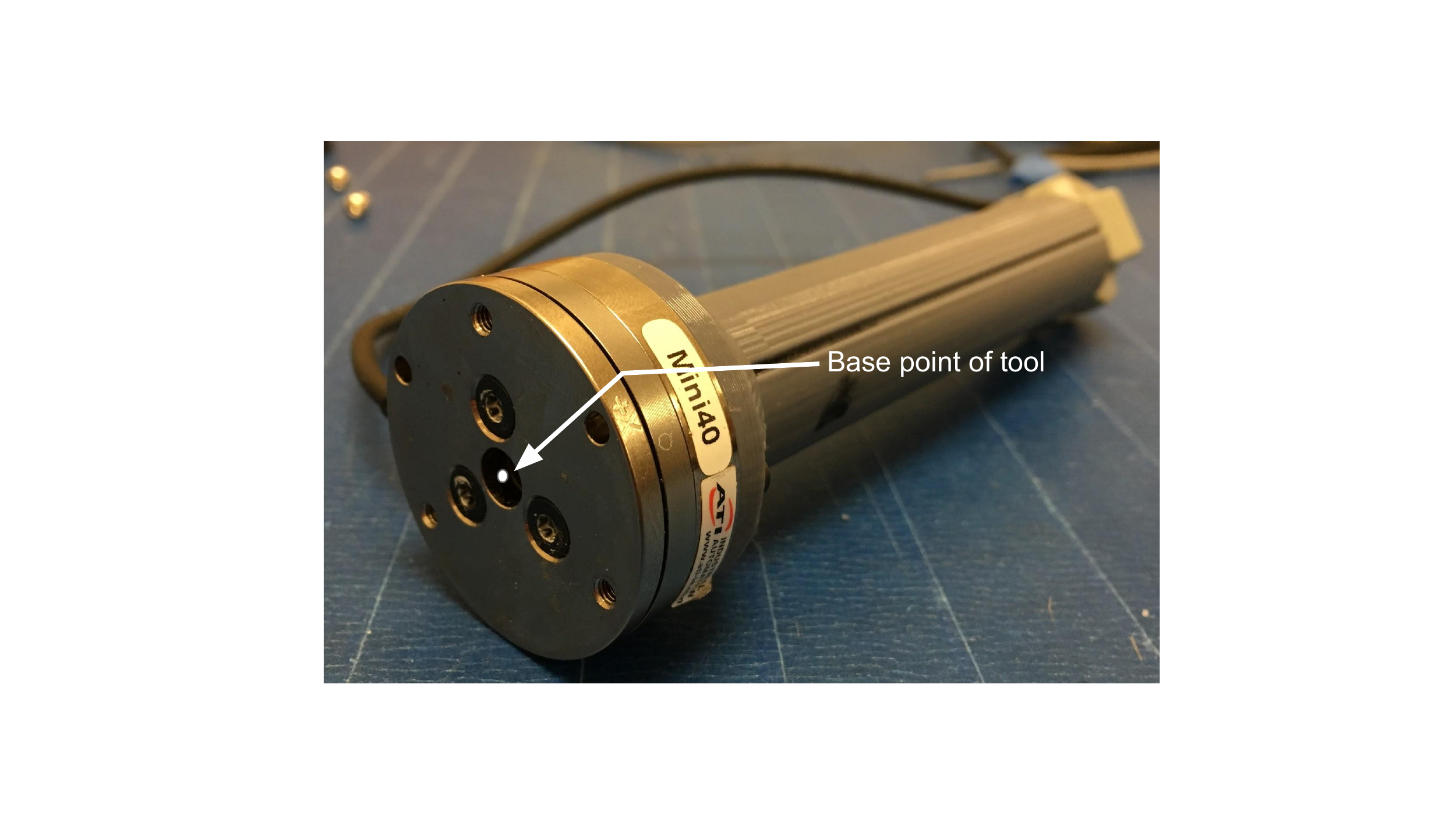}
		\caption{\add{The \emph{base point of the tool} is the center of the tooling side of the FT sensor}}
		\label{fig-base_point_tool}
	\end{figure}
	
\begin{figure}
		\centering
		\includegraphics[width=0.9\linewidth]{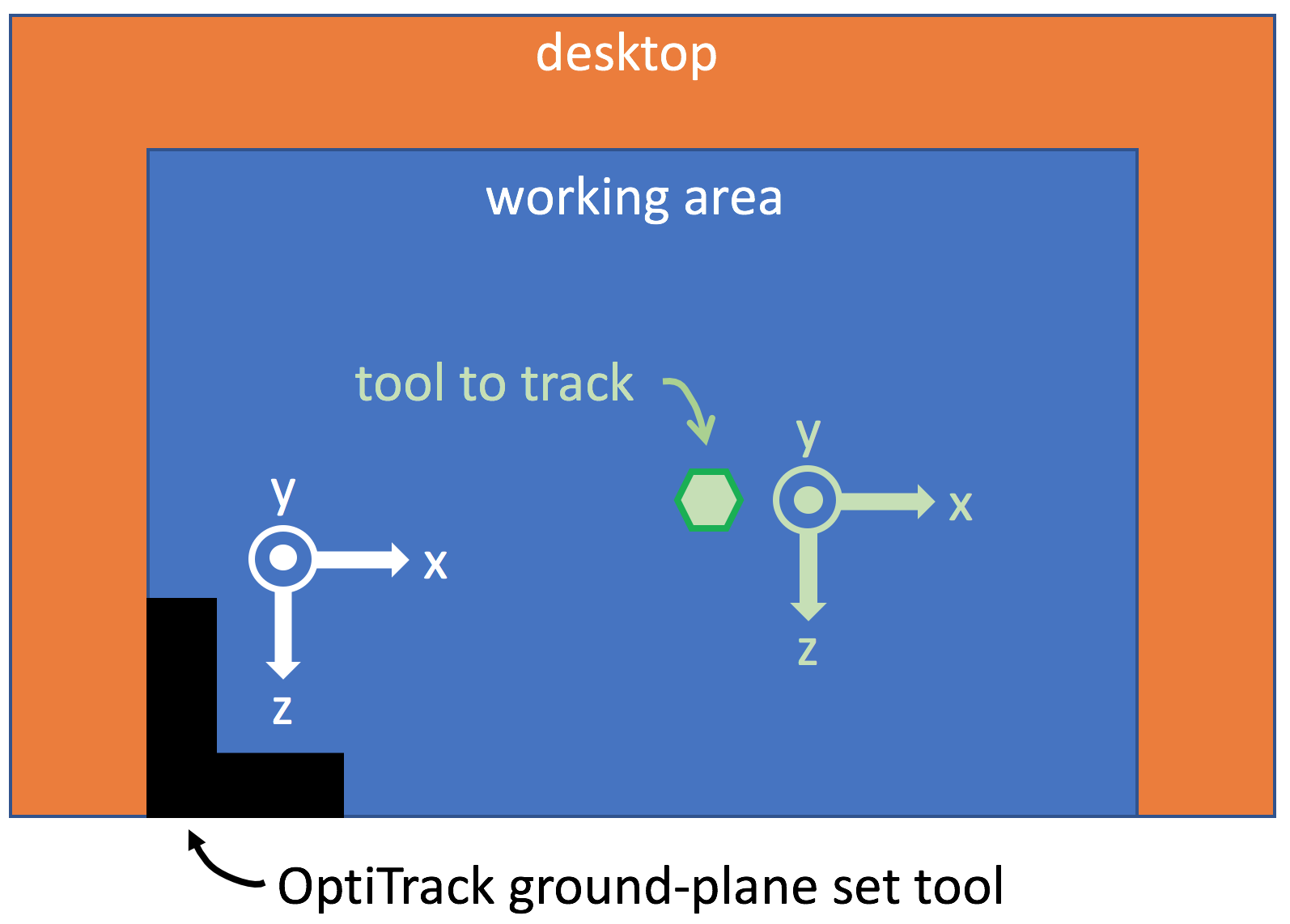}
		\caption{Top view of setting the coordinate frame of the ground plane and the trackable using OptiTrack.}
		\label{fig-ground_plane.png}
	\end{figure}

	To track a tool using OptiTrack, we need to define the ground plane and define the tool as a trackable. The ground plane is set by  \add{aligning} a right-angle set tool \add{to} the bottom left corner of the working area  The trackable is defined from a set of selected markers, and is assigned the same coordinate frame, with the origin being the centroid of the markers. This is shown in Fig. \ref{fig-ground_plane.png}. 
	
\begin{figure}
		\centering
		\includegraphics[width=0.7\linewidth]{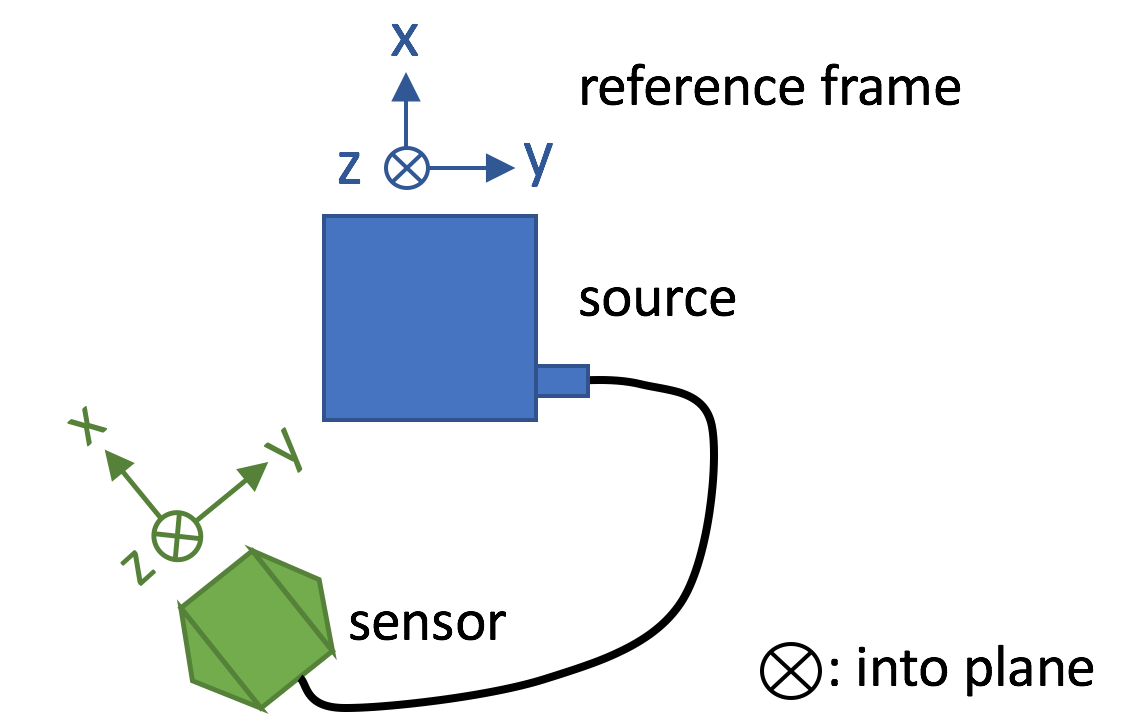}
		\caption{Illustration of the Patriot source and sensor when they are placed on the same plane, and the corresponding coordinate frames. $\bigotimes$ means \emph{into} the paper plane.}
		\label{fig-patriot_source_sensor}
	\end{figure}

	Patriot contains a source  that supports up to two sensors. The source provides the reference frame for the sensors \add{as shown in} Fig. \ref{fig-patriot_source_sensor}.  \add{We define the base point of the tool to be the center of the tooling side of the FT sensor, as shown in Fig. \ref{fig-base_point_tool}. The translation from the PO sensor to the base point of the tool is $[14.3, 0, 0.7]$, in the frame of the PO sensor, unit centimeter}. 
	
\begin{figure}
		\centering
		\includegraphics[width=0.9\linewidth]{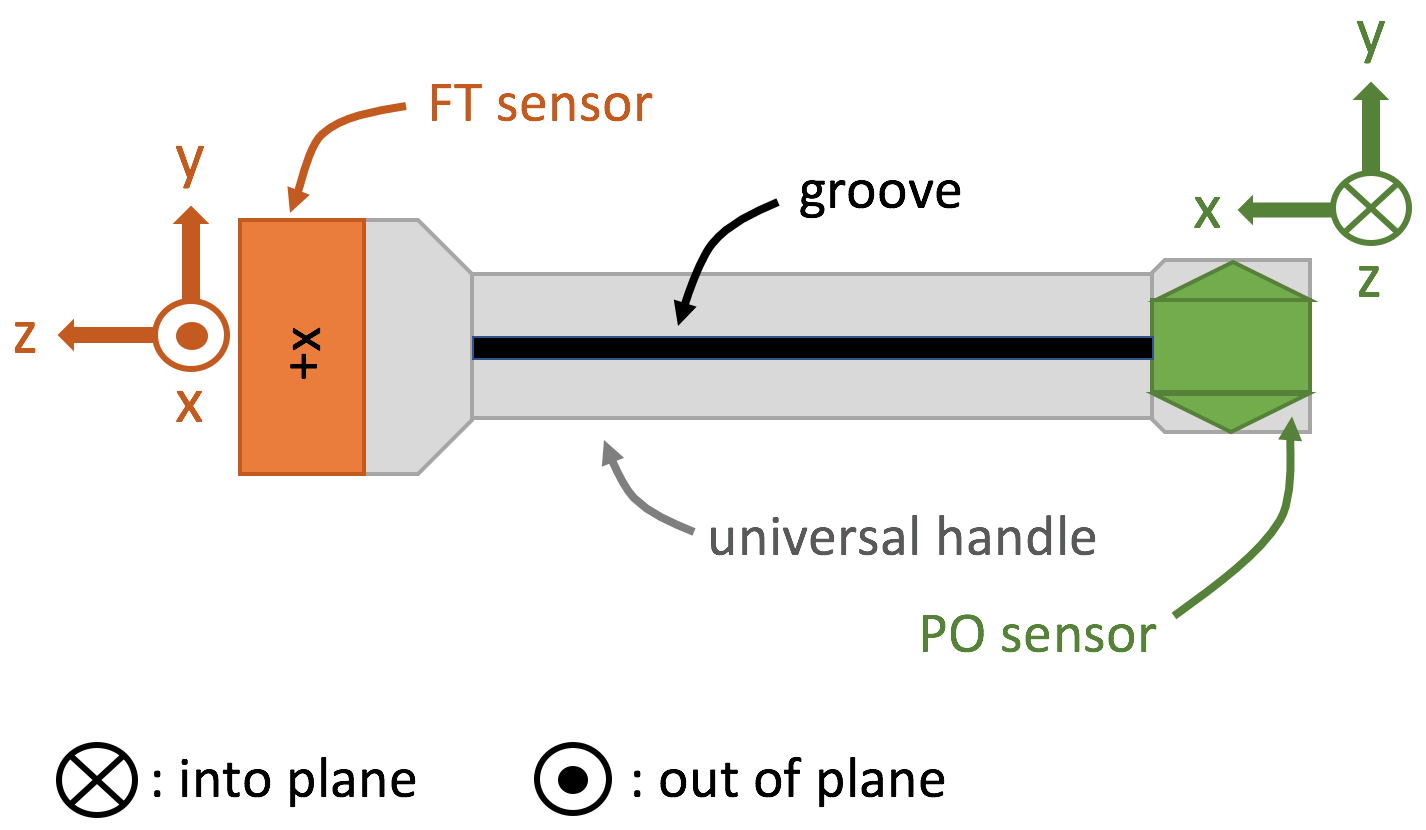}
		\caption{Top view of the FT sensor with its local frame, the universal handle, and the PO sensor with its local frame. $\bigotimes$ means \emph{into} the paper plane, and $\bigodot$ means \emph{out of} the paper plane.} 
		\label{fig-ft_handle_po}
	\end{figure}

	The FT sensor and the PO sensor are connected through the universal handle. The groove on the universal handle is orthogonal to both the $x-y$ plane of the FT sensor and the $y-z$ plane of the PO sensor. The relationship between the local frames of the FT sensor and the PO sensor is shown in Fig. \ref{fig-ft_handle_po}.
	
	\section{Calibrate FT}

	\begin{definition} \label{def-level}
		The level pose of the universal handle is a pose in which the groove of the handle faces up, and in which the $y-z$ plane of the FT sensor or equivalently the $x-y$ plane of the PO sensor is parallel to the desk surface. 
	\end{definition}
	
	\begin{definition} \label{def-average_reading}
		An average sample is the average of 500 FT samples. 
	\end{definition}
	The FT sensor has non-zero readings when it is static with the tool installed on it. We calibrate the FT sensor, or make the readings zeros, before we collect any data. We hold the handle in a level pose (Definition \ref{def-level}), and take an average sample (Definition \ref{def-average_reading}) which we set as the bias $FT_b$. We subtract the bias from each FT sample before saving the sample:
	$FT_t \leftarrow FT_t - FT_b.
	$     
	We calibrate the FT sensor each time we switch to a new tool.

	\section{Modality Synchronization}      
	
	Different modalities run at different frequencies and therefore need synchronization, which we achieve by using time stamps. We use Microsoft QueryPerformanceCounter (QPC) to query time stamps with millisecond precision. 
	
	When we start the collection system, we query the time stamp and set it as the global start time $t_0$. Then we start each modality as an independent thread, so that they run simultaneously and do not affect each other. For each sample, a modality queries the time stamp $t$ through QPC, and set the difference between $t$ and $t_0$, i.e. the elapsed time since $t_0$ as the time stamp for that sample:
	\begin{equation} \label{eq-time}
	t \leftarrow t - t_0.
	\end{equation} 
	
	\section{Data Format} \label{sec-label}
	
\begin{figure}
		\includegraphics[width=0.5\linewidth]{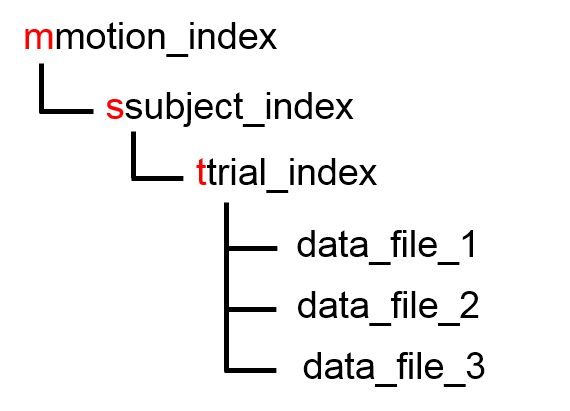}
		\caption{The structure of the dataset where the \textcolor{red}{red text} is verbatim.}
		\label{fig-hierarchy}
	\end{figure}

	The data are organized in a ``motion $\rightarrow$ subject $\rightarrow$ trial $\rightarrow$ data files" hierarchy, as shown in Fig. \ref{fig-hierarchy}, where the prefixes for motion, subject, and trial directories are \code{m}, \code{s}, and \code{t}, respectively. 
	
	RGB videos save as .\code{avi}, depth images save as .\code{png},  and the rest data files save as .\code{csv}. Both RGB and depth have a resolution of 640$\times$480, and are collected at 30Hz.  
	
\begin{figure}
		\includegraphics[width=\linewidth]{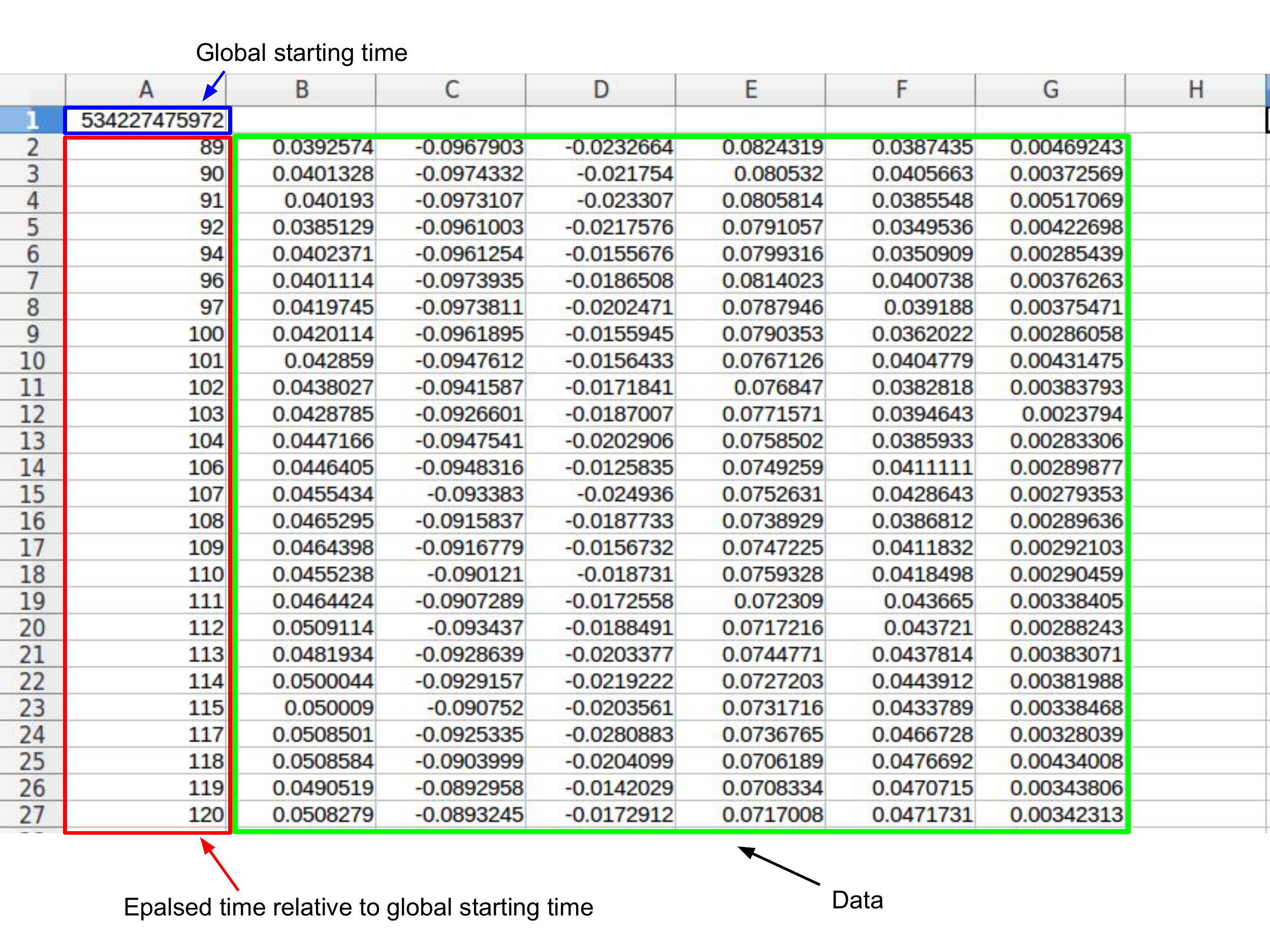}
		\caption{The structure of a non-OptiTrack \code{csv} data file. }
		\label{fig-data_example}
	\end{figure}

	
	The \code{csv} files excluding those of OptiTrack follow the same structure as shown in Fig. \ref{fig-data_example}. The first row contains the global start time and is the same in all the \code{csv} files that belong to the same trial. Starting with the second row, each row is a data sample, of which the first column is the time stamp (Eq. \eqref{eq-time}), and the rest columns are the data specific to a certain modality. The OptiTrack \code{csv} file differs in that it contains a single-column row between the start-time row and the data rows, which contains the number of defined trackables (1 or 2). In the following we explain the data part of a row for each different \code{csv} file. 
	
	FT sensor outputs 6 columns: ($f_x, f_y, f_z, \tau_x, \tau_y, \tau_z$), where $f_x$ and $\tau_x$ are the force and torque in the $+x$ direction, respectively. FT can be sampled at a very high frequency but we set it to be 1 kHz. The force has unit pound (lbf) and the torque has unit pount-foot (lbf-ft). 
	
	For the RGB videos and depth image sequences, we provide the time stamp for each frame in a \code{csv} file. The data part has one column, which is the frame index.

\begin{figure}
		\includegraphics[width=\linewidth]{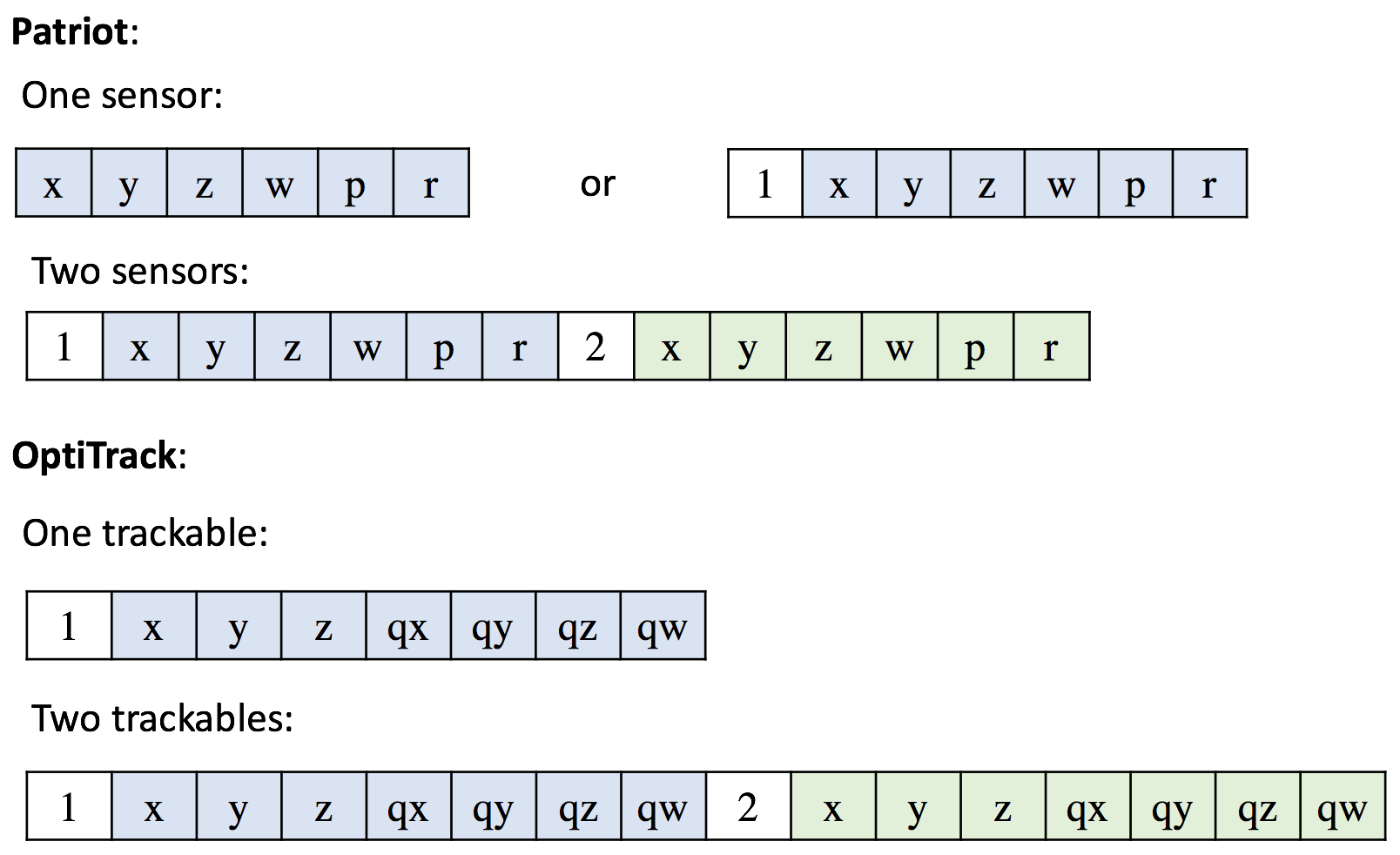}
		\caption{Formats of the columns for PO for one and two sensors}
		\label{fig-csv_contents}
	\end{figure}

\begin{figure}
		\centering
		\includegraphics[width=0.6\linewidth]{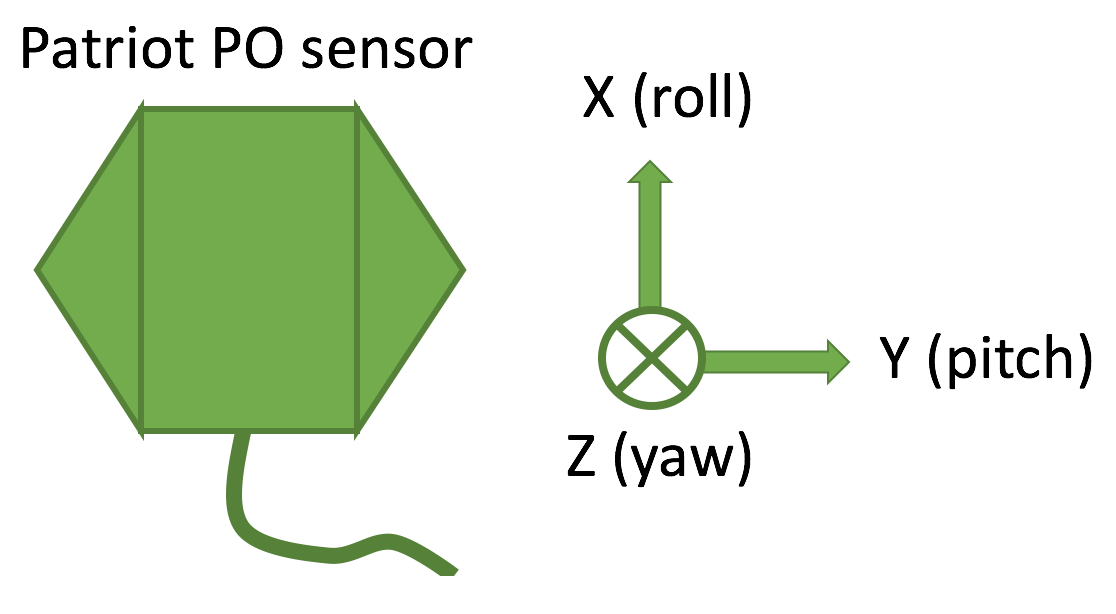}
		\caption{The relationship between the axes and yaw-pitch-roll for the Patriot sensor}
		\label{fig-patriot_ypr}
	\end{figure}

	The PO data contain the tool, and \emph{may} also contain the object. With two PO capture systems, and with or without the object, four different formats exist for the PO data, which are listed in Fig. \ref{fig-csv_contents}. Patriot expresses the orientation using yaw-pitch-roll (w-p-r) which is depicted in Fig. \ref{fig-patriot_ypr}, and OptiTrack uses unit quaternion (qx, qy, qz, qw). If we only use one trackable but have defined two in OptiTrack, we disable the inactive one by setting all 7 columns for that trackable to be -1, i.e., the 8 columns for the inactive trackable would be (1, -1, -1, -1, -1, -1, -1, -1).  
	
	Patriot samples at 60 Hz, its $x-y-z$ has unit centimeter and its yaw-pitch-roll has unit degree. OptiTrack samples at 100 Hz, and its $x-y-z$ has unit meter.  
	
	\section{Using the data}
	
\begin{figure}
		\includegraphics[width=\linewidth]{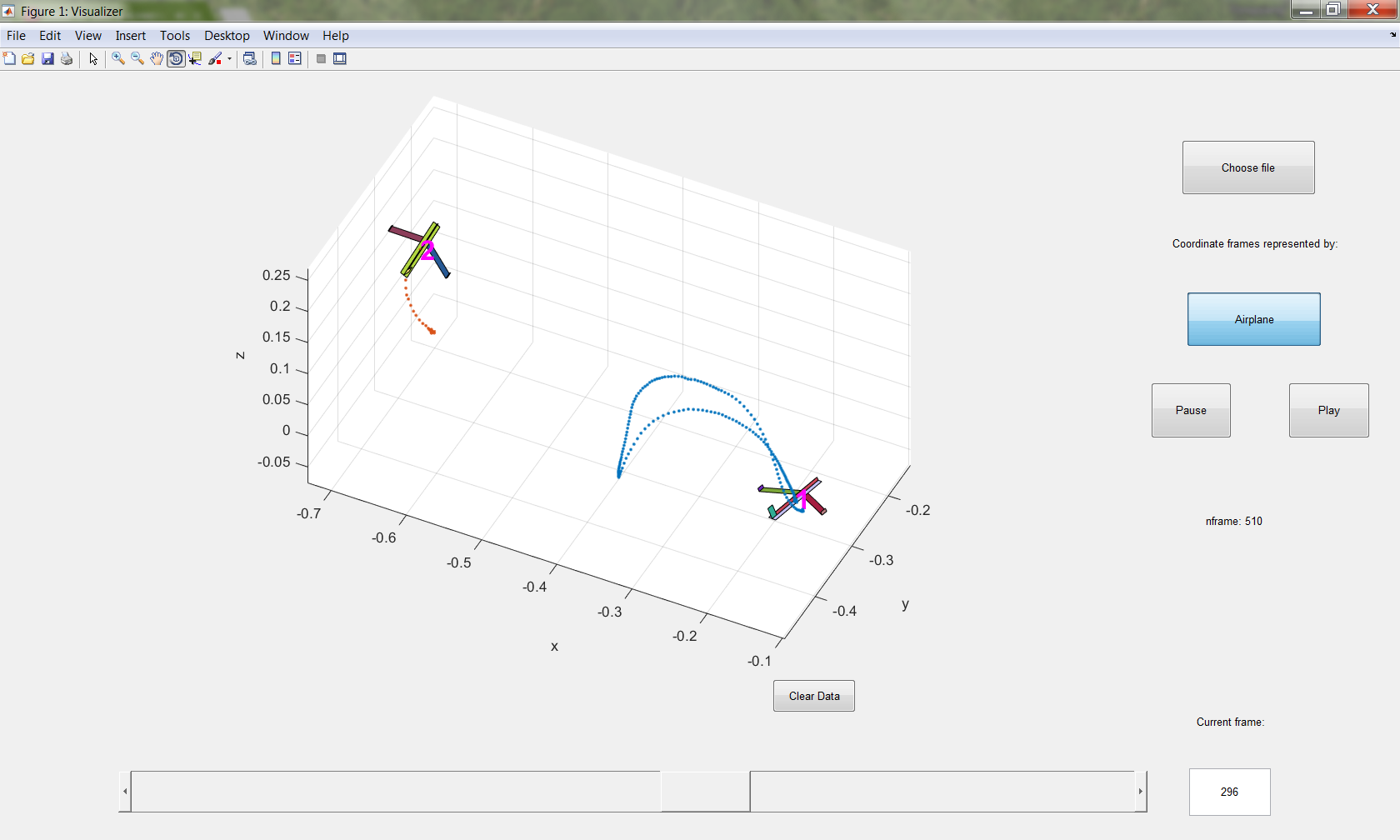}
		\caption{Visualizing the PO data}
		\label{fig-viz}
	\end{figure}

	We provide MATLAB code that visualizes the PO data for OptiTrack as well as Patriot, as shown in Fig. \ref{fig-viz}. The visualizer displays the trail of the \add{base point of the} tool \add{(Fig. \ref{fig-base_point_tool})} and the object if applicable as the motion is played as an animation in 3D. The user can also manually slide through the motion forward or backward and go to a particular frame.  
	
	The FT and PO \code{csv} files have multiple formats, and we provide Python code that extracts FT and PO data from each trial given the path of the root folder. Although we have explained the format of the \code{csv} files of the FT and PO data in Sec. \ref{sec-label}, we highly recommend using our code to get the FT and PO data to avoid error.  
	
	Each modality is sampled at a unique frequency, and using multiple modalities requires using the time stamps. One or more modalities need upsampling or downsampling.
	
	\section{Known issue}	
	
	The PO data recorded using OptiTrack contain occasional flickering and stagnant frames. This is caused by the dependency of OptiTrack on the line of sight. This issue is not present in the data collected with Patriot. 
	
	%

	\section{The pouring data} \label{sec-pouring}
	
	We want to learn to perform a type of motion from its PO and FT data, and generalize it, i.e., execute it in a different environment. Thus, we need data that show how the motion vary in multiple different environments. We realize that since pouring is the second frequently executed motion in cooking \cite{pauliusfunctional}, it is worth learning. Also, collecting pouring data that contain different environment setup is easy thanks to the convenience of switching material, cups, and containers. Therefore, we collected the pouring data.
	
	The pouring data include FT, Patriot PO, and RGB videos (no depth). We collected the data using the same system as described above. In the following, we explain what has not been covered and what differs from above. 
	
	The physical entities involved in a pouring motion include the material to be poured, the container from which the material is poured which we refer to as \emph{cup}, and the container to which the material is poured which we refer to as \emph{container}. The pouring data contain 1,596 trials of pouring water, ice, and beans from six different cups to ten different containers. Cups are considered as tools and are installed on the FT sensor through 3D-printed adapters.

	A second PO sensor is taped on the outer surface of the container just below the mouth.  
	
	We collect the FT data differently from above. When the cup is empty, we hold the handle in a level pose (Definition \ref{def-level}), and take an average sample (Definition \ref{def-average_reading}) which we call ``FT\_empty". Then we fill the cup with the material to an amount we desire, hold the handle in a level pose, and take an average sample which we call ``FT\_init". Then we pour, during which we take however many samples (\emph{not} average samples) which we call ``FT". After we finish pouring, we hold the handle in a level pose, and take an average sample which we call ``FT\_final". In summary, we save four kinds of FT data files -- three contain an average sample each: FT\_empty, the FT\_init, FT\_final, and one contains regular samples: FT. We do not consider bias.  
	
\begin{figure}
		\includegraphics[width=0.9\linewidth]{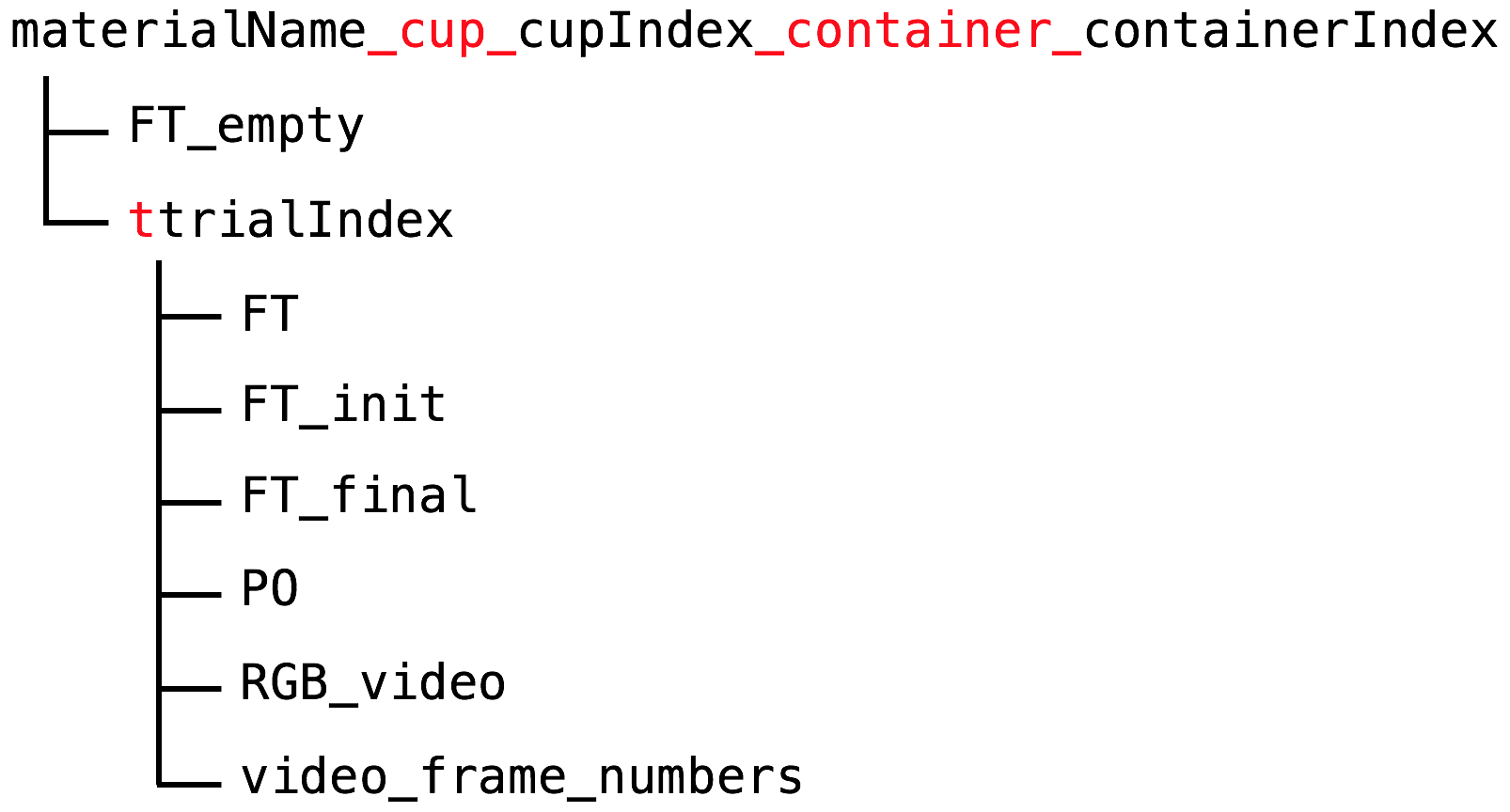}
		\caption{The organization of the pouring data where the \textcolor{red}{red text} is verbatim}
		\label{fig-pouring_org}
	\end{figure}

	The organization of the data is shown in Fig. \ref{fig-pouring_org}.     
	
	The pouring data can be used to learn how to pour in response to the sensed force of the cup. The force is a non-linear function of the physical properties of the cup and the material, the speed of pouring, the current pouring angle, the amount of remaining material in the cup, as well as other possibly related physical quantities. \cite{huang2017} shows an example of modeling such function using recurrent neural network and generalizing the pouring skills to unseen cups and containers.     
	
	\section{Conclusion \& Future work}
	
	We have presented a dataset of daily interactive manipulation. The dataset includes 32 types of motions, and provides position and orientation, and force and torque for every motion  \add{trial}. In addition, to support motion generalization to different environments, we chose the pouring motion and collected corresponding data. We plan to extend the collection to other types of motions in the future.

	\section{Acknowledgments}
	This material is based upon work supported by the National Science Foundation under Grants No. 1421418 and No. 1560761.    
	
	\bibliographystyle{SageH}
	\bibliography{motion_list}
	
\end{document}